\pdfoutput=1

\documentclass[11pt]{article}

\usepackage[final]{acl}

\usepackage{times}
\usepackage{latexsym}

\usepackage{graphicx} 
\usepackage{mathrsfs} 
\usepackage{booktabs}
\usepackage{amsmath} 
\usepackage{rotating} 
\usepackage{multirow} 
\usepackage[T1]{fontenc}

\usepackage[utf8]{inputenc}

\usepackage{microtype}

\usepackage{inconsolata}

\usepackage{graphicx}

\title{LogicPro: Improving Complex Logical Reasoning via Program-Guided Learning}

\author{
    \textbf{Jin Jiang\textsuperscript{1}}, 
    \textbf{Yuchen Yan\textsuperscript{2}}, 
    \textbf{Yang Liu\textsuperscript{3}}, 
    \textbf{Jianing Wang\textsuperscript{3}}, 
    \textbf{Shuai Peng\textsuperscript{1}}, 
\\
    \textbf{Xunliang Cai\textsuperscript{3}}, 
    \textbf{Yixin Cao\textsuperscript{4}}, 
    \textbf{Mengdi Zhang\textsuperscript{3}},     
    \textbf{Liangcai Gao\textsuperscript{1}\thanks{Corresponding author.}}
\\
\\
    \textsuperscript{1}Wangxuan Institute of Computer Technology, Peking University, Beijing, China, 
\\
    \textsuperscript{2}College of Computer Science and Technology, Zhejiang University, 
    \textsuperscript{3}Meituan Group, 
\\
    \textsuperscript{4}Institute of Trustworthy Embodied AI, Fudan University
\\
\small{
    \textbf{Correspondence:} \href{mailto:jiangjin@stu.pku.edu.cn}{jiangjin@stu.pku.edu.cn}, 
    \href{mailto:gaoliangcai@pku.edu.cn}{gaoliangcai@pku.edu.cn}
}
}

\begin{document}
\maketitle
\begin{abstract}
In this paper,  we propose a new data synthesis method called \textbf{LogicPro}, which leverages LeetCode-style algorithm  \underline{Pro}blems and their corresponding \underline{Pro}gram solutions to synthesize Complex \underline{Logic}al Reasoning data in text format.
First, we synthesize complex reasoning problems through source algorithm problems and test cases.
Then, standard answers and intermediate variable outputs are obtained for each problem based on standard python solutions and test cases.
Finally, with the guidance of code intermediate variables, we synthesize the text reasoning process for each reasoning problems.
Through this method, we can synthesize data that is difficult, scalable, effective, and comes with golden standard answers and high-quality reasoning processes.
As a result, with our 540K synthesized dataset constructed solely from 2,360 algorithm problems, our approach \footnote{Code and data are publicly available at \url{https://github.com/jiangjin1999/LogicPro}} achieves significant improvements in multiple models for the datasets \textit{BBH$^{27}$}, \textit{LogicBench}, \textit{DROP}, \textit{AR-LSAT}, and \textit{GSM8K}, etc. outperforming a wide range of existing reasoning datasets.
\end{abstract}

\section{Introduction}

With the rapid development of artificial intelligence, Large Language Models (LLMs) \citep{bi2024deepseek, liu2024deepseek} demonstrate excellent performance in reasoning tasks. The success of these models is inseparable from the support of large-scale and high-quality reasoning data. However, data acquisition and processing face numerous challenges in the real world. As a viable alternative, synthetic data \citep{wang2024survey} can effectively alleviate this problem and further enhance \citep{dubey2024llama, adler2024nemotron} the model's reasoning capabilities.
\begin{figure}[h]
    \centering
    \includegraphics[width=0.47\textwidth]{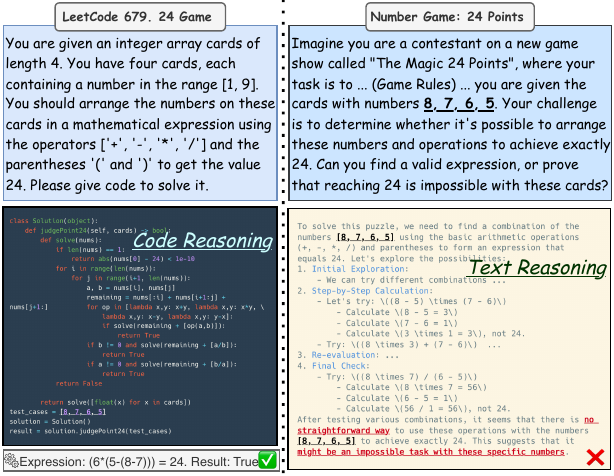}
    \caption{Left: LeetCode 679. 24 Game - original algorithm problem and standard Python solution. Right: Our synthesized complex reasoning problem: 24-point and g4o-api response.}
    \label{fig:intro}
\end{figure}

Synthetic data \citep{liu2024best} has wide applications in mathematics and code domains.
For mathematical data, synthetic data can be generated through problem-driven methods such as evol-instruct \citep{luo2023wizardmath, zeng2024automatic}, problem restatement \citep{yumetamath}, back-translation \citep{lu2024mathgenie}, and few-shot examples \citep{dalvi2021explaining}, or knowledge-driven methods based on knowledge bases \citep{dalvi2021explaining} and concept graphs \citep{tangmathscale}.
For code data, it generates diverse and high-quality instruction data with self-evolutionary methods \citep{wang2023self, luowizardcoder} or combined with open-source references such as OSS-Instruct \citep{wei2024magicoder}.

In the complex logical reasoning domain, there is relatively less research on data synthesis. Previously, some work uses logical proposition-based soft reasoning methods \citep{tafjord2021proofwriter, clark2021transformers} to synthesize training data, aiming to improve models' reasoning ability and interpretability \citep{saeed2021rulebert, dalvi2021explaining}.
Additionally, \citet{sinha2019clutrr} and \citet{tian2021diagnosing} use kinship graphs and first-order logic to synthesize person relationship reasoning data and logical entailment reasoning data respectively.
More recently, \citet{morishita2023learning} using formal logic theory, \citet{parmar2024logicbench} covering 25 patterns across different logic types, and \citet{morishitaenhancing} utilizing formal language programs to synthesise reasoning data.

Previous methods primarily used propositional logic or formal languages as the source of logic.
Instead, we find that algorithmic questions and programming languages like LeetCode provide a different source of logic. 
Algorithmic problems typically involve explicit input-output relationships, recursive and iterative structures, and operations on data structures. All of these constitute a unique pattern of reasoning.
At the same time, algorithmic problems are naturally related to real-world task contexts, such as path-planning, data sorting, and resource allocation problems, which can often be mapped to real-world application scenarios.

As shown in Figure \ref{fig:intro}, we discover that while an algorithm problem can be easily solved through code, LLMs often make mistakes when the same problem is transformed into a specific text reasoning question. 
Inspired by this observation, we propose a data synthesis method aimed at generating high-quality reasoning problems and reasoning processes by utilizing widely available algorithmic problems and their code solutions.

Our approach consists of data collection and three steps. For data collection, we collect a large number of existing LeetCode algorithmic problems and their code solutions, while collecting or constructing diverse test cases.
First, by combining the original algorithm problem with test cases, it is transformed into a specific text reasoning problem. Then, the original standard code solution is combined with the test cases to generate the corresponding python code solution. By running the code, the final output result and the values of key intermediate variables can be obtained. Finally, based on the code output, the model is guided to synthesize the complete reasoning process for the current problem.

This data synthesis method offers significant advantages: sufficient difficulty, scalability, effectiveness, and high-quality reasoning paths. 
Specifically, the sufficient difficulty is reflected in models performing worse on LogicPro compared to other baseline data; scalability was demonstrated by collecting more algorithmic problems and constructing more test cases to further scale the data; effectiveness is shown through performance improvements across multiple models on multiple Out-of-Distribution (OOD) benchmarks. 

Our main contributions are summarized as follows:
\begin{enumerate}
    \item We propose a novel data synthesis method called LogicPro, which uses LeetCode-style data as seeds to synthesize text-formatted complex logical data through algorithmic problems and program solution.
    \item With this approach, we can synthesise a 540K dataset from just 2,360 algorithmic problems that is sufficiently difficult, scalable, and effective, as well as having standard answer and high-quality reasoning process.
    \item The experimental results show that our approach achieves significant improvements in multiple models for the datasets \textit{BBH$^{27}$}, \textit{LogicBench}, \textit{DROP}, \textit{AR-LSAT}, and \textit{GSM8K}, etc. outperforming a wide range of existing datasets.
\end{enumerate}

\begin{figure*}[htp]
    \centering
    \includegraphics[width=1\textwidth]{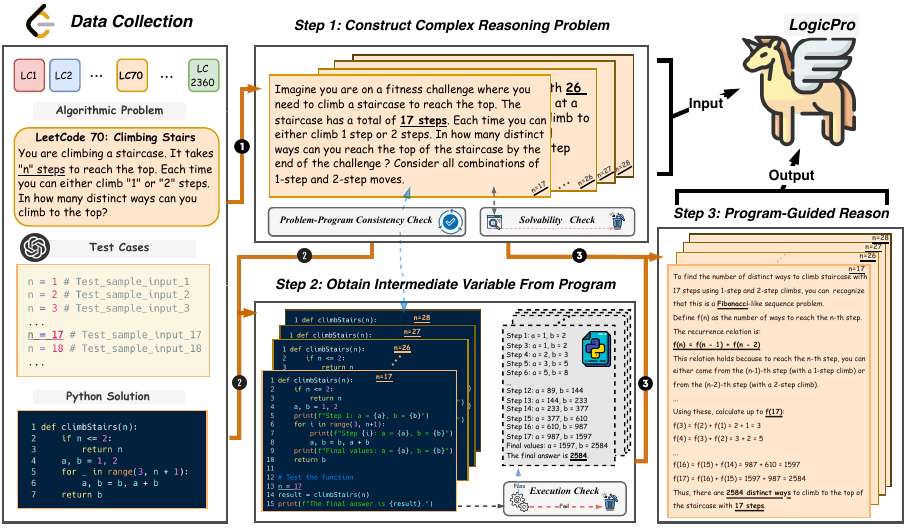}
    \caption{An overview of LogicPro (Example: LeetCode-70 Climbing Stairs): 1. Construct Complex Reason Problem (Section \ref{sec:step1} Step 1), based on source algorithm problems and test cases to synthesize complex reasoning problems. 2. Obtain Intermediate Variable From Program (Section \ref{sec:step2} Step 2) 3. Program-Guided Reason (Section \ref{sec:step3} Step 3), synthesizing the final Input-Output from complex reasoning problems and intermediate variable outputs.}
    \label{fig:main_method}
\end{figure*}

\section{Approach}
\label{sec:Approach}

In this section, we elaborate on the LogicPro data synthesis process.
It mainly includes: data collection and construction of test cases (Section \ref{sec:data_collection}), how to synthesize textual complex logical reasoning problems (Section \ref{sec:step1}), how to obtain intermediate variable outputs from code (Section \ref{sec:step2}), and how to synthesize high-quality reasoning processes based on intermediate variable outputs and synthesized questions (Section \ref{sec:step3}). An overview of our approach is shown in Figure \ref{fig:main_method}.

\subsection{Data Collection}
\label{sec:data_collection}

In the data collection phase, source Leet data was collected test cases were constructed. We collect 2,360 official LeetCode problems as initial seeds for LogicPro's synthetic data. 
However, due to the limited number of test cases in the original data, GPT-4 is used to reconstruct test cases for each question.
Specifically, we prompt GPT-4 to generate 150 test cases for each LeetCode problem, sample three times, and then perform consolidation, deduplication, and filtering on the three results. 
Eventually, 2,360 LeetCode algorithm problems were compiled with standard Python solutions, each containing up to 300 test cases.

\subsection{Step 1: Construct Complex Reasoning Problem}
\label{sec:step1}

\begin{figure}[h]
    \centering
    \includegraphics[width=0.4\textwidth]{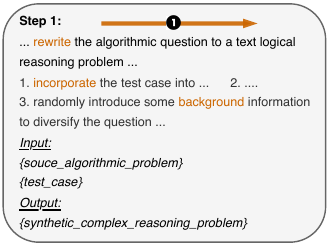}
    \caption{Schematic Diagram of Step 1.}
    \label{fig:step1}
\end{figure}

In the first step, we synthesize complex reasoning problems based on the original algorithm problems and test cases. As shown in Figure \ref{fig:step1}, the input includes the source LeetCode algorithm problems and specific test case. In this process, the LLM needs to combine test cases and algorithm problems, while randomly adding some background information, to synthesize a complex reasoning problem. Referring to the example in Figure \ref{fig:main_method}, the model combine LeetCode Problem 70 "Climbing Stairs" and the test case input n=17 into a specific text reasoning problem.

For the data flow in this step, 699K complex reasoning problems in text format are generated. Subsequently, after conducting consistency check (5K+) and solvability check, we select 595K qualified problems for the next step.

\paragraph{Problem-Program Consisitency Check}
The purpose of this module is to examine the consistency between code and text reasoning problem. 
In the second step, we rewrite the original standard Python solution and generate test case specific python code. 
To ensure consistency between the rewritten Python code and the text reasoning problems generated in the first step, we perform consistency checks on the code and problems. 
After inspection, 5K+ data entries are filtered out. Prompt is in Appendix \ref{sec:modules} Figure \ref{fig:Consisitency}.

\paragraph{Solvability Check}
This check module is designed to check whether the synthesised questions are solvable.
In the first step, a large number of textual reasoning problems were generated by combining the original LeetCode problems with the test cases generated by GPT-4. 
However, some of these problems are unsolvable or meaningless. This is mainly because the test cases generated by GPT-4 are not entirely perfect. Although GPT-4 ensures that the generated test cases comply with the problem requirements and code format, these test cases may still deviate from the core testing points of the problems, or the sample length is too long, resulting in unsolvable synthesized problems. After detection, we have screened out and filtered 98K unsolvable problems. Prompt is in Appendix \ref{sec:modules} Figure \ref{fig:Solvability}.

\subsection{Step2: Obtain Intermediate Variable from Program}
\label{sec:step2}

\begin{figure}[htp]
    \centering
    \includegraphics[width=0.4\textwidth]{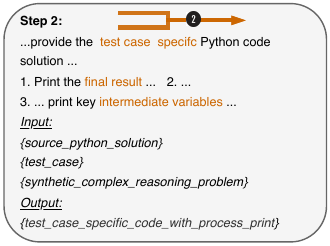}
    \caption{Schematic Diagram of Step 2.}
    \label{fig:step2}
\end{figure}
In the second step, we input Python code solutions, test cases, and synthesized reasoning problems to generate test case specific python code. As shown in Figure \ref{fig:step2}, we require the model to generate python code solutions related to the test cases and ensure that the code prints the final result to the result variable. 
In addition, to simulate the human reasoning process, the model is required to print key intermediate variables in the code.

For example, for the climbing stairs problem shown in Figure \ref{fig:main_method}, the model needs to generate python code for the specific test case of climbing 17 stairs, and print intermediate variables in the code, namely the calculation process of the Fibonacci sequence at each step. 
Finally, by executing the code, the output of the intermediate variables can be obtained, which will assist in the reasoning synthesis for the next step. 

For the data flow of this step, the final answer to each problem is obtained by running the combined code that integrates the test cases with the standard Python solution.
Meanwhile, we modify the code to print important intermediate variable values. Then, through execution check, we finally obtain 544K data points, including standard answers and code intermediate variable outputs for each problem.

\paragraph{Execution Check}
This check module is designed to detect errors that occur during code execution.
In the step 2, the output of intermediate variables in the code was obtained by modifying the original standard Python solution
for these errors, we perform data filtering, which can be divided into two categories: The first category involves modified code that encounters errors during execution and still fails to run properly after multiple sampling attempts; The second category involves code that executes correctly but still contains error messages in the intermediate variable outputs due to the use of try-catch functionality. After detection, these two types of issues result in the filtering of 50K error codes. 

\subsection{Step3: Program-Guided Reasoning}
\label{sec:step3}

\begin{figure}[htp]
    \centering
    \includegraphics[width=0.4\textwidth]{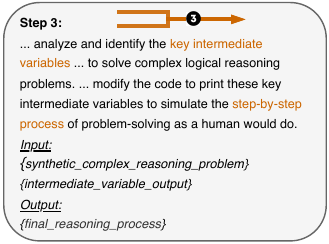}
    \caption{Schematic Diagram of Step 3.}
    \label{fig:step3}
\end{figure}

In the third step, the input consists of the complex reasoning problem generated in the first step and the intermediate variable output obtained in the second step. These elements are used to synthesize the final reasoning process.
As shown in Figure \ref{fig:step3}, the model (Llama3.1-70B-Instruct) answers synthesized complex reasoning questions by analyzing intermediate variable data.

Taking the example in Figure \ref{fig:main_method}, based on the calculation process of the Fibonacci equation in the intermediate variable output, the model successfully synthesized a high-quality reasoning process as the final output of LogicPro data. The goal of this step is to generate golden reasoning processes with logic and accuracy, thereby ensuring the completeness and credibility of the reasoning.

For the data flow in this step, based on the text-format complex reasoning problems and corresponding code intermediate variable outputs obtained from previous steps, we use this important intermediate variable information to guide the model in generating high-quality reasoning processes for complex reasoning problems. 
Finally, a high-quality resoning quetion-answer datasets with 540k input-output pairs is obtained.

\section{Experiments}

\subsection{Experimental Setup}

\subsubsection{Training}
This section elaborates on the experimental settings related to training. We introduce: (1) the baseline datasets used for comparison with our LogicPro;  (2) experimental design to verify the effect of different synthetic data. (3) the model configurations used in training and analysis; (4) the parameters and implementation details of the training process.

\paragraph{Baseline Data}
To comprehensively evaluate the effectiveness of our method, we select multiple mainstream logical reasoning data synthesis methods as baseline comparisons, including RuleTakers \citep{clark2021transformers}, LogicNLI \citep{tian2021diagnosing}, ProofWriter \citep{tafjord2021proofwriter}, CLUTRR \citep{sinha2019clutrr}, RuleBert \citep{saeed2021rulebert}, LogicBench \citep{parmar2024logicbench}, and FLD \citep{morishita2023learning}. The detailed information of these datasets can be found in Appendix \ref{sec:Baseline_data_statistics} Table \ref{tab:Baseline_data_statistics}.
\paragraph{Experimental Design}
To better simulate the actual training scenarios of LLMs, we use two types of base data: first, 100K general domain data extracted from OpenHermes-2.5 \citep{OpenHermes}, and second, LogiCoT \citep{liu2023logicot} as the specialized base data for logical reasoning domain. By mixing these different base data with baseline data and our synthetic data for training separately, we systematically analysed the impact of different synthetic data on model performance.

\paragraph{Models}
We conduct systematic experimental research on foundation models of various scales and architectures. The main experiments employ four representative models: Qwen-7B and Llama3-8B as small-scale representatives, and Qwen2-72B and Llama3-70B as large-scale model representatives. For in-depth understanding of model behaviors, we select Qwen2-7B and Llama-8B for fine-grained analysis experiments. 
\paragraph{Training Details}
In terms of training implementation, we use Megatron-LM as the training framework with the following configurations: a cosine learning rate schedule is adopted with an initial learning rate of 1e-5, a warmup ratio of 0.03, and the learning rate decays to 0; the maximum sequence length is set to 8192, with a global batch size of 128, and the number of training epochs is set to 3. All experiments are completed with Supervised Fine-tuning (SFT) on a computing cluster consisting of 32 NVIDIA A100 GPUs.

\subsubsection{Evaluation}
This section details the experimental setup associated with the evaluation. We present: (1) the benchmarks used to evaluate the different synthetic data; and (2) the implementation details of the evaluation process.

\paragraph{Benchmarks}
To comprehensively evaluate the model's complex logical reasoning capabilities, ten representative benchmark datasets for testing are selected , including BBH$^{27}$ \citep{suzgun2023challenging}, LogicBench \citep{parmar2024logicbench}, DROP \citep{dua2019drop}, AR-LSAT \citep{zhong2021ar}, Boardgamqa \citep{kazemi2024boardgameqa}, FOLIO \citep{han2022folio}, GSM8K \citep{cobbe2021training}, Multi-LogicEval \citep{patel2024multi}, ProofWriter \citep{tafjord2021proofwriter}, and MATH \citep{hendrycks2measuring}.
Notably, for our synthetic data LogicPro, all benchmark are out-of-domain tests. In comparison, some baseline data, same as some benchmark, are synthesized based on propositional logic or first-order logic. Additionally, LogicBench's training and test sets are completely in-domain.
Furthermore, BBH, as a core benchmark for evaluating models' complex logical reasoning capabilities, includes 27 challenging reasoning tasks. Based on this, we apply weights to all data subsets to calculate the final average score.
Due to space limitations, only the evaluation results of seven benchmarks are presented in the main text, with complete results available in Appendix \ref{sec:full_result} Table \ref{tab:full_result}.

\paragraph{Evaluation Details}
In the inference phase, we use the vLLM \citep{kwon2023efficient} framework for deployment. The inference configuration adopts greedy decoding strategy and sets the maximum generation length to 4096 tokens.
For the evaluation of model output, we adopt Qwen-2.5-72B as the model evaluator to score. The specific evaluation prompt template can be found in Figure \ref{fig:model_eval}.

\subsection{Main Results}
\begin{table*}[t]
  \centering
  \resizebox{\linewidth}{!}{%
    \begin{tabular}{lcccccccc}
    \toprule
    \multicolumn{1}{l}{\textbf{Model}} & \textbf{BBH$^{27}$} & \textbf{LogicBench} & \textbf{DROP} & \textbf{ARLSAT} & \textbf{\#BoardQA} & \textbf{FOLIO} & \textbf{GSM8K} & \textbf{\#Avg} \\
    \midrule
    Qwen2-7B-RuleTakers & 45.4 & 59.1 & 65.7 & 16.5 & 42.2 & 44.6 & 80.9 & 45.8 \\
    Qwen2-7B-LogicNLI & 43.3 & 71.3 & 67.4 & \underline{17.8} & 45.3 & 41.7 & 81.6 & 45.0 \\
    Qwen2-7B-ProofWriter & 40.8 & 68.6 & 64.3 & 17.0 & 36.9 & 36.3 & 80.9 & 43.2 \\
    Qwen2-7B-CLUTRR & 43.0 & \underline{72.0} & 64.0 & 17.0 & \textbf{51.9} & 41.2 & 80.4 & 45.0 \\
    Qwen2-7B-RuleBert & \underline{46.2} & 69.1 & 67.4 & 16.5 & 43.8 & 40.7 & \underline{81.7} & \underline{47.3} \\
    Qwen2-7B-LogicBench & 44.7 & *95.9 & 67.4 & \underline{17.8} & 41.4 & 38.7 & \textbf{82.1} & 46.6 \\
    Qwen2-7B-FLD & 42.0 & 69.5 & \textbf{68.3} & 14.8 & 34.9 & \underline{45.6} & 80.0 & 43.8 \\
    \midrule
    Qwen2-7B-LogicPro (ours) & \textbf{50.9} & \textbf{73.5} & \textbf{68.3} & \textbf{19.1} & \underline{48.1} & \textbf{46.1} & 81.5 & \textbf{51.2} \\
    \midrule
    \midrule
    Llama3-8B-RuleTakers & 38.5 & 59.9 & 65.9 & 12.6 & \textbf{47.3} & 43.3 & 67.9 & 40.3 \\
    Llama3-8B-LogicNLI & 40.4 & 54.0 & 65.3 & 12.6 & 41.0 & \underline{56.0} & 69.3 & 41.8 \\
    Llama3-8B-ProofWriter & 37.2 & 62.1 & 66.4 & \textbf{15.2} & 31.3 & 47.8 & 69.1 & 39.5 \\
    Llama3-8B-CLUTRR & 40.5 & 61.1 & \underline{66.6} & 10.4 & 43.7 & \underline{56.0} & 69.5 & 42.2 \\
    Llama3-8B-RuleBert & 34.7 & 48.8 & 66.5 & \textbf{15.2} & 43.6 & 51.5 & 68.9 & 37.4 \\
    Llama3-8B-LogicBench & \underline{41.0} & *93.5 & 66.2 & 10.9 & 38.6 & \textbf{62.7} & \underline{69.8} & \underline{43.7} \\
    Llama3-8B-FLD & 35.7 & \underline{67.8} & 61.2 & \underline{13.5} & 39.1 & 50.0 & 64.6 & 38.2 \\
    \midrule
    Llama3-8B-LogicPro (ours) & \textbf{45.0} & \textbf{67.9} & \textbf{68.8} & \textbf{15.2} & \underline{44.3} & 48.3 & \textbf{74.2} & \textbf{46.2} \\
    \midrule
    \midrule
    Qwen2-72B-RuleTakers & 61.3 & 72.4 & 76.6 & 19.6 & \underline{61.0} & 49.3 & 88.5 & 61.3 \\
    Qwen2-72B-LogicNLI & 61.7 & \underline{80.7} & 77.0 & 21.3 & 60.4 & 58.2 & 87.3 & 61.6 \\
    Qwen2-72B-ProofWriter & 61.8 & 75.5 & 77.2 & 16.5 & 55.0 & 44.0 & 88.0 & 61.5 \\
    Qwen2-72B-CLUTRR & \underline{68.1} & 79.0 & \underline{78.4} & 24.4 & \underline{61.0} & 59.7 & \textbf{89.4} & 66.7 \\
    Qwen2-72B-RuleBert & 67.8 & 74.1 & 76.5 & 19.6 & 55.2 & \textbf{62.7} & 88.0 & 66.5 \\
    Qwen2-72B-LogicBench & 67.1 & *97.0 & 77.9 & \underline{24.8} & 57.3 & \underline{60.5} & 88.4 & \underline{67.2} \\
    Qwen2-72B-FLD & 65.4 & 72.4 & 76.3 & 17.0 & 45.5 & 53.7 & 86.9 & 63.6 \\
    \midrule
    Qwen2-72B-LogicPro (ours) & \textbf{72.4} & \textbf{81.7} & \textbf{79.6} & \textbf{27.4} & \textbf{66.4} & 55.2 & \underline{89.1} & \textbf{70.4} \\
    \midrule
    \midrule
    Llama3-70B-RuleTakers & 51.5 & 69.1 & 75.6 & 19.1 & 61.5 & 53.9 & \underline{86.8} & 52.7 \\
    Llama3-70B-LogicNLI & 58.5 & 69.9 & 78.0 & 17.8 & 58.5 & \underline{58.3} & 84.5 & 57.8 \\
    Llama3-70B-ProofWriter & 55.3 & 31.9 & 75.3 & 15.2 & 58.7 & 51.0 & 65.4 & 53.2 \\
    Llama3-70B-CLUTRR & 57.8 & 71.8 & 74.0 & 20.9 & \underline{62.1} & \textbf{61.3} & 75.1 & 57.3 \\
    Llama3-70B-RuleBert & 56.0 & 68.7 & 75.0 & 13.9 & 51.5 & 51.0 & 85.4 & 55.1 \\
    Llama3-70B-LogicBench & \underline{61.4} & *93.2 & \underline{78.4} & \underline{21.3} & 58.7 & 50.0 & 84.6 & \underline{60.8} \\
    Llama3-70B-FLD & 57.2 & \underline{74.4} & 75.0 & 16.5 & 46.3 & 56.9 & 85.5 & 56.4 \\
    \midrule
    Llama3-70B-LogicPro (ours) & \textbf{63.7} & \textbf{72.7} & \textbf{78.8} & \textbf{22.3} & \textbf{65.0} & 54.2 & \textbf{87.6} & \textbf{62.4} \\
    \bottomrule
    \end{tabular}%
    }
    \caption{Main results of LogicPro with baseline data. Where \#BoardQA represents BoardgameQA  and \#Avg represents Average. \textbf{Bold} denotes the best score in that baseline, \underline{underline} denotes the second highest score, and * denotes same-distribution data. See Table \ref{tab:full_result} for more details.}
  \label{tab:main_result}%
\end{table*}%

Table \ref{tab:main_result} shows the main results where LogicPro outperforms previous synthesis methods across multiple benchmarks.
On the representative Big Bench Hard benchmark, LogicPro improves the average performance by 2.3\% - 4.7\% compared to the previous best baseline across different model type and scales, and shows at least 10\% improvement over general baseline data. On LogicBench (a benchmark measuring propositional logic capabilities), LogicPro achieves the best performance except when LogicBench itself is used as the training set. On FOLIO (a benchmark measuring first-order logic capabilities), except for Qwen2-7B, the performance of the other three models trained with LogicPro is inferior to other baseline data. This may be because some baseline synthetic data is essentially generated based on first-order logic, making their data distribution closer to FOLIO, leading to better performance. On GSM8K (mathematical reasoning benchmark), different synthetic data has relatively minor impact on model performance. On OOD benchmarks such as DROP (reading comprehension reasoning) and ARLSAT (law school admission test reasoning), LogicPro also demonstrates advantages across multiple model foundations, further validating its performance on out-of-distribution tasks.
The weighted average results across all benchmarks show that on the Qwen2 foundation, 7B and 72B models improve by 3.9\%-8\% and 3.1\%-9.1\% respectively compared to baselines; on the Llama3 foundation, 8B and 70B models improve by 2.5\%-8.8\% and 1.6\%-9.7\% respectively compared to baselines.

Additionally, for different scale models, we find that our LogicPro still shows advantages on large-scale models. 
This indicates that although large-scale models possess stronger reasoning capabilities, our data still provides important value for improving their performance.
It is speculated that this may be related to the high difficulty level of LogicPro itself, with detailed analysis in section \ref{sec:difficulty_comparison}.

\section{Analysis}

\subsection{Ablation Study}
    
\begin{table}[htbp]
  \centering
  \resizebox{\linewidth}{!}{%
    \begin{tabular}{cccc}
    \toprule
    \textbf{Qwen2-7B} & \textbf{BBH$^{27}$} & \textbf{LogicBench} & \textbf{\#Avg} \\
    \midrule
    Source\_LeetCode & 41.0 & 63.4 & 42.2  \\
    LogicPro$_{\text{w/o Inter-Var}}$ & 48.2$_{+7.2}$ & 69.6$_{+6.2}$ & 49.0$_{+6.8}$  \\
    \textbf{LogicPro$_{\text{w. Inter-Var}}$} & \textbf{50.9}$_{+9.9}$  & \textbf{73.5}$_{+10.1}$  & \textbf{51.2}$_{+9.0}$  \\
    \midrule
    \textbf{Llama3-8B} & \textbf{BBH$^{27}$} & \textbf{LogicBench} & \textbf{\#Avg} \\
    \midrule
    Source\_LeetCode & 36.6 & 51.4 & 38.5  \\
    LogicPro$_{\text{w/o Inter-Var}}$ & 44.0$_{+7.4}$ & 63.6$_{+12.2}$ & 45.1$_{+6.6}$  \\
    \textbf{LogicPro$_{\text{w. Inter-Var}}$} & \textbf{45.0}$_{+8.4}$  & \textbf{67.9}$_{+16.5}$  & \textbf{46.2}$_{+7.7}$  \\
    \bottomrule
    \end{tabular}%
    }
  \caption{Ablation study. Source\_LeetCode is the source 2360 LeetCode algorithm questions and code solution. LogicPro$_{\text{w/o Inter-Var}}$ and LogicPro$_{\text{w. Inter-Var}}$ indicate whether the construction process uses intermediate variables. In other words, LogicPro$_{\text{w/o Inter-Var}}$ is a direct distillation of the text problems in LogicPro and LogicPro$_{\text{w. Inter-Var}}$ is our final LogicPro data. Llama3.1-70B-Instruct is the model to distill and generate final reasoning trajectories.}
  \label{tab:ablation_study}%
\end{table}%

As shown in Table \ref{tab:ablation_study}, we conduct ablation studies to validate LogicPro's effectiveness in synthesizing questions and generating high-quality reasoning processes.
First, we compare Source\_LeetCode with LogicPro$_{\text{w/o Inter-Var}}$ to analyze the effect on synthesizing text reasoning questions. For BBH, our method improves the performance of Qwen2-7B and Llama3-8B by 7.21\% and 7.34\% respectively, with significant improvements of 6.8\% and 6.6\% on the Average metric. 
Subsequently, we compare LogicPro$_{\text{w/o Inter-Var}}$ with LogicPro$_{\text{w. Inter-Var}}$ to analyze the effect of code intermediate variables in the reasoning process. The results show that introducing code intermediate variables as guidance can further improve data quality compared to directly distilling synthesized questions. Specifically, compared to LogicPro$_{\text{w/o Inter-Var}}$, it achieves improvements of 2.68\% and 1.04\% on BBH, and improvements of 2.2\% and 1.1\% on the Average metric.

\subsection{Performance Gain on All Baseline Data}

\begin{table}[htbp]
  \centering
  \resizebox{\linewidth}{!}{%
    \begin{tabular}{cccc}
    \toprule
    \textbf{Qwen2-7B} & \textbf{BBH$^{27}$} & \textbf{LogicBench} & \textbf{\#Avg} \\
    \midrule
    All\_Baseline & 43.1  & 92.8  & 47.1 \\
    All* + LogicPro & \textbf{48.8}$_{+5.7}$  & \textbf{96.3}$_{+3.5}$  & \textbf{52.3}$_{+5.2}$ \\
    \midrule
    \textbf{Llama3-8B} & \textbf{BBH$^{27}$} & \textbf{LogicBench} & \textbf{\#Avg} \\
    \midrule
    All\_Baseline & 42.9  & 94.3  & 47.6 \\
    All* + LogicPro & \textbf{47.6}$_{+4.7}$  & \textbf{95.0}$_{+0.7}$  & \textbf{51.2}$_{+3.6}$ \\
    \bottomrule
    \end{tabular}%
    }
  \caption{Results of continuous performance improvement on existing data. All\_Baseline represents that we mix all baseline data from the main experiment. All* + LogicPro represents that we further mix LogicPro data for comparison.}
  \label{tab:performance_gain}%
\end{table}

As shown in Table \ref{tab:performance_gain}, we combine all baseline data to evaluate the effect of introducing LogicPro. The experimental results show that on top of integrating all baseline data, the addition of LogicPro can further improve model performance. On the BBH task, it brings improvements of 5.6\% and 4.7\% respectively, while for average performance, it improves by 5.2\% and 3.6\% respectively.

It is worth emphasizing that our goal is not to propose a method to replace existing synthetic data, but rather to introduce a novel data synthesis strategy. Different from previous synthetic data, our data's logic comes from LeetCode algorithm problems and programming languages, rather than propositional logic and formal languages. As the results above, our data complements existing data to further enhance the model's complex reasoning capabilities.

\begin{table*}[ht]
  \centering
  \resizebox{\linewidth}{!}{%
    \begin{tabular}{cccccccc|c}
    \toprule
    \textbf{Model} & \textbf{\#Rule} & \textbf{LogicNLI} & \textbf{\#Proof} & \textbf{CLUTRR} & \textbf{RuleBert} & \textbf{LogicBench} & \textbf{FLD} & \textbf{LogicPro} \\
    \midrule
    \textbf{Qwen2-7B-Instruct} & 73.0  & 62.8  & 64.5  & 46.3  & 53.6  & 74.5  & 54.1  & \textbf{39.3 } \\
    \textbf{Qwen2-72B-Instruct} & 78.3  & 78.9  & 73.8  & 72.0  & 54.1  & 77.3  & 69.1  & \textbf{46.3 } \\
    \textbf{Llama3-8B-Instruct} & 69.0  & 63.3  & 62.1  & 67.0  & 47.8  & 75.5  & 51.5  & \textbf{41.2 } \\
    \textbf{Llama3-70B-Instruct} & 80.6  & 79.6  & 78.9  & 83.8  & 65.5  & 82.8  & 64.4  & \textbf{55.6 } \\
    \midrule
    \textbf{Average} & 75.2  & 71.2  & 69.8  & 67.3  & 55.2  & 77.5  & 59.8  & \textbf{45.6 } \\
    \bottomrule
    \end{tabular}%
    }
  \caption{Results from different baseline data and LogicPro's difficulty comparisons on four open source models. \#Rule represents RuleTakers and \#Proof represents ProofWriter. The metrics for the results in the table are accuracy rates, and smaller is better.}
  \label{tab:Difficulty_Comparison}%
\end{table*}%

\subsection{Data Scaling Analysis}
\begin{figure}[h]
    \centering
    \includegraphics[width=0.5\textwidth]{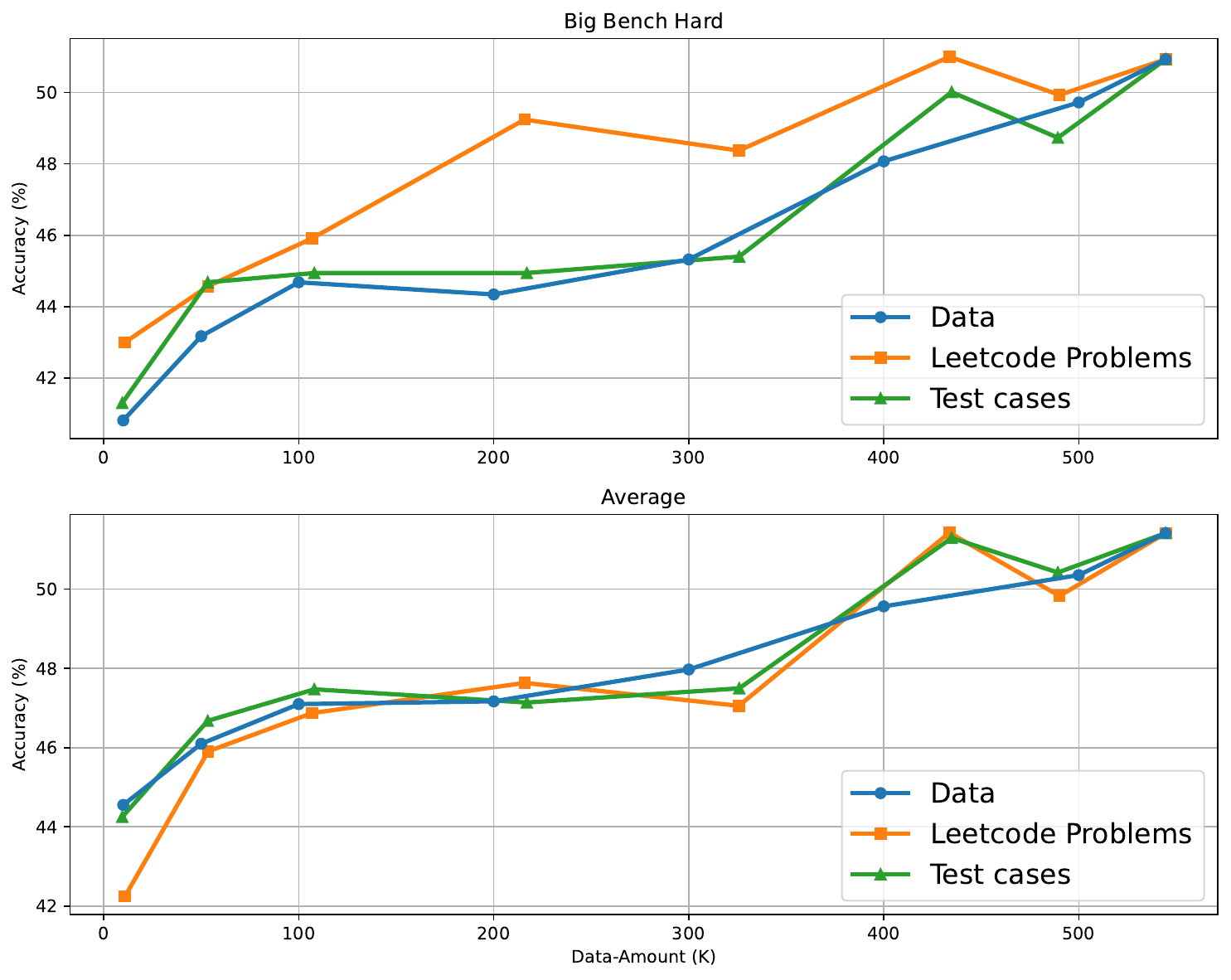}
    \caption{The results of the three scaling methods, \textit{Data}, \textit{Leetcode Problems} and \textit{Test Cases}, on BBH and Average. \textit{Data} means we randomly sample 10K-50K-100K$\sim$500K-540K from LogicPro; \textit{Leetcode Problems} means that we sample 2\%-10\%-20\%$\sim$90\%-100\% according to the number of algorithmic problems; \textit{Test cases} means that we select all algorithmic problems and sample the number of test cases for each problem according to 2\%-10\%-20\%$\sim$90\%-100\%.}
    \label{fig:data_scaling}
\end{figure}

In Figure \ref{fig:data_scaling}, we demonstrate the impact of three data scaling methods on BBH and Average metrics to analyze how the total data amount, number of algorithm problems, and number of test cases in data synthesis methods affect model performance.
Overall, the \textit{Data} scaling method shows a steady upward trend, indicating that the synthesized data after mixing effectively improves model performance, while also revealing potential for further improvement through expanded data scale.

In contrast, the \textit{Leetcode Problems} scaling method shows more fluctuation. As the number of Leetcode problems increases from 30\% to 100\%, both BBH and Average metrics display multiple patterns of decline followed by increase. This may be due to distributional differences in the data generated from different algorithm problems, leading to fluctuations in the mixed data results.

The \textit{Test Cases} method generally shows an upward trend, but model performance declines after the number of test cases increases to 80\%. This may be attributed to two factors: First, there is diminishing marginal utility of test cases. With a fixed number of algorithm problems, simply increasing test cases may reach a performance improvement bottleneck, and only further increasing the number of algorithm problems can effectively breakthrough. Second, the impact of test case quality: The large number of test cases directly generated by GPT4 are limited in quality. Although increasing quantity brings some improvement, constructing higher-quality test cases is still needed for more significant optimization. 

In summary, to further improve the performance of the model, it is necessary to enhance the quality of test cases while increasing the number of algorithmic problems. This can break through the bottleneck of data expansion. On this basis, the model's capabilities can be continuously enhanced.

\subsection{Difficulty Comparison with Baseline Data}
\label{sec:difficulty_comparison}


As mentioned above, LogicPro is a sufficiently difficult dataset.
To further validate this point, in Table \ref{tab:Difficulty_Comparison} we compare the performance  between various baseline datasets and LogicPro across four open-source models. Specifically, we randomly sample 5K samples from each baseline dataset and LogicPro for evaluation, where LogicPro uses a uniform sampling strategy to ensure each algorithm problem is fairly selected.

The results show that four open-source models Qwen2-7B-Instruct, Qwen2-72B-Instruct, Llama3-8B-Instruct, and Llama3-70B-Instruct perform quite well on existing baseline datasets, with average accuracy exceeding 50\%, and some datasets even reaching over 70\%. However, their performance drops significantly on the LogicPro dataset, with an average accuracy of only 45.6\%. Notably, as mentioned in the "Main Results" section, LogicPro still poses significant challenges for Large Language Models, which may explain why there is still considerable space for improvement in their performance on large-scale models.


\section{Related work}
\subsection{Synthetic Reasoning Data}
Synthetic data has played an important role in enhancing LLMs' reasoning capabilities \citep{dubey2024llama}, especially in mathematics and coding domains.

For mathematical reasoning, synthetic data generation includes problem-driven methods such as evol-instruct \citep{luo2023wizardmath, zeng2024automatic}, problem restatement \citep{yumetamath}, back-translation \citep{lu2024mathgenie}, and few-shot examples \citep{dalvi2021explaining}, as well as knowledge-driven methods that rely on knowledge bases \citep{dalvi2021explaining} or concept graphs \citep{tangmathscale} to generate new problems by sampling key reasoning points \citep{huang2024key}.

In terms of code reasoning \citep{chen2025revisit}, from Code Alpaca's \citep{wang2023self} use of self-instruct to generate 20K instruction data based on seed tasks, to WizardCoder's \citep{luowizardcoder} use of Code evol-instruct to generate more complex tasks, to Magicoder's use of oss-instruct \citep{wei2024magicoder} to extract 75K instructions from open source code. Synthesizing data continuously improves the model's code reasoning capabilities.

In contrast, there is less research on synthetic data for complex logical reasoning. Learning combines formal logic theory \citep{morishita2023learning} to synthesize data using basic reasoning rules to train language models' multi-step reasoning abilities. LogicBench \citep{parmar2024logicbench} not only constructs logical reasoning benchmark datasets but also provides synthetic data based on formal logic.

LeetCode-style algorithm problems contain rich reasoning processes. This paper synthesizes high-quality complex logical  sreasoning data based on the formal logic of programming languages to enhance models' reasoning capabilities.

\subsection{Symbolic Reason Enhances LLM Reason}
Symbolic language was initially used for formal logical reasoning, mathematical computation, and program verification, playing a crucial role in early artificial intelligence applications such as expert systems \citep{hatzilygeroudis2004neuro} and automated theorem provers \citep{loveland2016automated}. With the rapid development of LLMs, symbolic reasoning, as a structured reasoning approach, has further enhanced the reasoning capabilities of large models through their integration.

There are two main approaches to combining symbolic reasoning with LLMs. The first approach utilizes symbolic language for planning and reasoning. Some research combines LLMs with symbolic solvers like Python \citep{gao2023pal} and SAT \citep{ye2023satlm} to solve mathematical \citep{he23solving} and logical reasoning problems \citep{pan2023logic}, reducing the reasoning burden on models. Additionally, some work employs symbolic language as a planning \citep{hao2024planning, wen2024codeplan} tool to strengthen LLMs' reasoning capabilities.

The second approach involves using symbolic language to generate training data. Early research \citep{tafjord2021proofwriter, clark2021transformers} adopted neural network-based soft reasoning methods to synthesize training data from logical rules expressed in natural language, improving reasoning capabilities and interpretability \citep{saeed2021rulebert, dalvi2021explaining}. Recently, more research has explored the application of symbolic methods in synthetic training data. \citet{lineuro} formalizes seed data and replaces certain variables to synthesize mathematical reasoning data; \citet{morishita2023learning} constructs synthetic corpora based on formal logic; \citet{shao2024case2code} generates inductive reasoning data through case-to-code mapping; \citet{parmar2024logicbench} provides more comprehensive propositional logic benchmarks; \citet{morishitaenhancing} strengthens LLM reasoning capabilities through principled synthesis of logical corpora. These studies have advanced the development of logical reasoning capabilities.

In this paper, we further explore the application of symbolic methods in data synthesis, utilizing formal logical information embedded in LeetCode algorithm problems to synthesize complex logical reasoning data.

\section{Conclusion and Future Work}

This paper presents a new data synthesis method called LogicPro. This approach utilizes LeetCode-style algorithm problems and solutions to generate complex logical reasoning data.
By synthesizing a 540K dataset from just 2,360 seed problems, our approach ensures scalability, difficulty, and high-quality reasoning paths. Results show that LogicPro enhances model performance across multiple reasoning benchmarks, including BBH$^{27}$, LogicBench, DROP, AR-LSAT, and GSM8K, outperforming a wide range of existing datasets. 

For future work, considering the vast number of algorithmic problems in the real world, such as problems from Luogu, ACM competitions, and various online judges, we can collect more algorithmic problem data to further expand LogicPro's dataset. Additionally, our data includes both result signals (code execution results as \textbf{standard answers}) and process signals (intermediate variables during code execution), which may provide new insights for reinforcement learning.
\section*{Limitations}

Our method explores an algorithm based on LeetCode-style approach to synthesize complex logical reasoning data.
In the future, one of our improvement directions is to build more comprehensive and diverse test cases. However, the generation of test cases itself is an independent research field. More advanced test generation techniques are expected to further enhance the quality and generalization ability of synthetic datasets.
Furthermore, although LeetCode's official 2,360 problems have achieved significant results, there are still numerous high-quality algorithm problems in the real world, such as Luogu, ACM competitions, and various Online Judge (OJ) platforms. Meanwhile, new algorithm problems continue to emerge. 
If these resources can be fully utilized, the quality and coverage of synthetic data will be further enhanced, leading to a larger data scale and better performance.

\section*{Ethics Statement}

This study is based on data from 2360 algorithmic questions on the fully open-source LeetCode platform. All data are from publicly available sources and do not involve any personal privacy information. Our study strictly adheres to the terms of use and privacy policies of the platforms from which the data was sourced. We ensure that the rights of all users and platform regulations are respected during data collection and processing. Through the use of publicly available data, we aim to advance academic research and education, and promote progress in the field of algorithms and computer science

\section*{Acknowledgments}

This work is supported by the projects of Beijing Science and Technology Program (Z231100007423011) and  National Natural Science Foundation of China (No. 62376012), which is also a research achievement of State Key Laboratory of Multimedia Information Processing and Key Laboratory of Science, Technology and Standard in Press Industry (Key Laboratory of Intelligent Press Media Technology).

\bibliography{acl_latex}


\appendix



\section{Details of Experiment}

\subsection{Baseline Data Statistics}
\label{sec:Baseline_data_statistics}

In Table \ref{tab:Baseline_data_statistics}, we statistics on the size of the baseline data, the logical sources of the synthetic data.

\begin{table}[htbp]
  \centering
  \resizebox{\linewidth}{!}{
    \begin{tabular}{llll}
    \toprule
    \textbf{Dataset} & \textbf{Size} & \textbf{Logic Source} & \textbf{Source} \\
    \midrule
    RuleTakers & 480K  & Soft Rule & \citep{clark2021transformers} \\
    LogicNLI & 48K   & FOL   & \citep{tian2021diagnosing} \\
    ProofWriter & 580K  & Soft Rule (proof) & \citep{tafjord2021proofwriter} \\
    CLUTRR & 50K   & Kinship Graph & \citep{sinha2019clutrr} \\
    RuleBert & 310K  & Soft Rule & \citep{saeed2021rulebert} \\
    LogicBench & 12K   & Propositional Logic & \citep{parmar2024logicbench} \\
    FLD   & 300K  & Formal Logic & \citep{morishita2023learning} \\
    \midrule
    LogicPro (ours) & 540K  & Program Logic & LogicPro \\
    \bottomrule
    \end{tabular}%
    }
    \caption{Data Statistics.}
  \label{tab:Baseline_data_statistics}%
\end{table}%

\subsection{Complete Main Results}
\label{sec:full_result}
Given space constraints, we show the complete results for all benchmarks in Table \ref{tab:full_result}.

\begin{table*}[htbp]
  \centering
  
  \resizebox{\linewidth}{!}{%
    \begin{tabular}{lccccccccccc}
    \toprule
    \multicolumn{1}{l}{\textbf{Model}} & \textbf{BBH$^{27}$} & \textbf{LogicBench} & \textbf{DROP} & \textbf{AR-LSAT} & \textbf{BoardgameQA} & \textbf{FOLIO} & \textbf{GSM8K} & \textbf{Multi-LogiEval} & \textbf{ProofWriter} & \textbf{MATH} & \textbf{Average} \\
    \midrule
    Qwen2-7B-RuleTakers & 45.4 & 59.1 & 65.7 & 16.5 & 42.2 & 44.6 & 80.9 & 53.8 & 20.0 & 40.8 & 45.8 \\
    Qwen2-7B-LogicNLI & 43.3 & 71.3 & 67.4 & 17.8 & 45.3 & 41.7 & 81.6 & 48.7 & 36.5 & 40.6 & 45.0 \\
    Qwen2-7B-ProofWriter & 40.8 & 68.6 & 64.3 & 17.0 & 36.9 & 36.3 & 80.9 & 43.4 & 63.5 & 40.3 & 43.2 \\
    Qwen2-7B-CLUTRR & 43.0 & 72.0 & 64.0 & 17.0 & 51.9 & 41.2 & 80.4 & 56.3 & 36.9 & 40.1 & 45.0 \\
    Qwen2-7B-RuleBert & 46.2 & 69.1 & 67.3 & 17.8 & 43.8 & 40.7 & 81.7 & 59.3 & 34.6 & 40.7 & 47.3 \\
    Qwen2-7B-LogicBench & 44.7 & 95.9* & 67.4 & 17.8 & 41.4 & 38.7 & 82.1 & 67.3 & 18.0 & 41.3 & 46.6 \\
    Qwen2-7B-FLD & 42.0 & 69.5 & 68.3 & 14.8 & 34.9 & 45.6 & 80.0 & 50.5 & 37.3 & 40.6 & 43.8 \\
    \midrule
    Qwen2-7B-LogicPro (ours) & 50.9 & 73.5 & 68.3 & 19.1 & 48.1 & 46.1 & 81.5 & 62.0 & 28.5 & 41.6 & 51.2 \\
    \midrule
    \midrule
    Llama3-8B-RuleTakers & 38.5 & 59.9 & 65.9 & 12.6 & 47.3 & 43.3 & 67.9 & 64.2 & 29.4 & 19.8 & 40.3 \\
    Llama3-8B-LogicNLI & 40.4 & 54.0 & 65.3 & 12.6 & 41.0 & 56.0 & 69.3 & 62.3 & 35.0 & 19.8 & 41.8 \\
    Llama3-8B-ProofWriter & 37.2 & 62.1 & 66.4 & 15.2 & 31.3 & 47.8 & 69.1 & 62.8 & 45.4 & 19.5 & 39.5 \\
    Llama3-8B-CLUTRR & 40.5 & 61.1 & 66.6 & 10.4 & 43.7 & 56.0 & 69.5 & 63.1 & 35.1 & 20.0 & 42.2 \\
    Llama3-8B-RuleBert & 34.7 & 48.8 & 66.5 & 15.2 & 43.6 & 51.5 & 68.9 & 65.4 & 28.2 & 19.5 & 37.4 \\
    Llama3-8B-LogicBench & 41.0 & 93.5* & 66.2 & 10.9 & 38.6 & 62.7 & 69.8 & 65.2 & 38.9 & 20.2 & 43.7 \\
    Llama3-8B-FLD & 35.7 & 67.8 & 61.2 & 13.5 & 39.1 & 50.0 & 64.6 & 60.0 & 38.4 & 17.8 & 38.2 \\
    \midrule
    Llama3-8B-LogicPro (ours) & 45.0 & 67.9 & 68.8 & 15.2 & 44.3 & 48.3 & 74.2 & 68.0 & 37.7 & 23.0 & 46.2 \\
    \midrule
    \midrule
    Qwen2-72B-RuleTakers & 61.3 & 72.4 & 76.6 & 19.6 & 61.0 & 49.3 & 88.5 & 60.7 & 70.6 & 54.6 & 61.3 \\
    Qwen2-72B-LogicNLI & 61.7 & 80.7 & 77.0 & 21.3 & 60.4 & 58.2 & 87.3 & 64.3 & 48.9 & 53.6 & 61.6 \\
    Qwen2-72B-ProofWriter & 61.8 & 75.5 & 77.2 & 16.5 & 55.0 & 44.0 & 88.0 & 63.0 & 70.7 & 55.0 & 61.5 \\
    Qwen2-72B-CLUTRR & 68.1 & 79.0 & 78.4 & 24.4 & 61.0 & 59.7 & 89.4 & 59.2 & 54.0 & 56.2 & 66.7 \\
    Qwen2-72B-RuleBert & 67.8 & 74.1 & 76.5 & 19.6 & 55.2 & 62.7 & 88.0 & 71.7 & 60.2 & 55.2 & 66.5 \\
    Qwen2-72B-LogicBench & 67.1 & 97.0* & 77.9 & 24.8 & 57.3 & 60.5 & 88.4 & 87.9 & 57.2 & 55.4 & 67.2 \\
    Qwen2-72B-FLD & 65.4 & 72.4 & 76.3 & 17.0 & 45.5 & 53.7 & 86.9 & 60.0 & 56.5 & 54.7 & 63.6 \\
    \midrule
    Qwen2-72B-LogicPro (ours) & 72.4 & 81.7 & 79.6 & 27.4 & 66.4 & 55.2 & 89.1 & 72.1 & 54.3 & 55.8 & 70.4 \\
    \midrule
    \midrule
    Llama3-70B-RuleTakers & 51.5 & 69.1 & 75.6 & 19.1 & 61.5 & 53.9 & 86.8 & 74.4 & 33.1 & 36.3 & 52.7 \\
    Llama3-70B-LogicNLI & 58.5 & 69.9 & 78.0 & 17.8 & 58.5 & 58.3 & 84.5 & 67.5 & 31.3 & 34.5 & 57.8 \\
    Llama3-70B-ProofWriter & 55.3 & 31.9 & 75.3 & 15.2 & 58.7 & 51.0 & 65.4 & 50.3 & 49.3 & 24.6 & 53.2 \\
    \midrule
    Llama3-70B-LogicPro (ours) & 63.7 & 72.7 & 78.8 & 22.3 & 65.0 & 54.2 & 87.6 & 72.3 & 33.5 & 40.5 & 62.4 \\
    \bottomrule
    \end{tabular}%
    }
    \caption{Complete evaluation results for all benchmark.}
  \label{tab:full_result}%
\end{table*}%

\begin{figure*}[htp]
    \centering
    \includegraphics[width=1.0\textwidth]{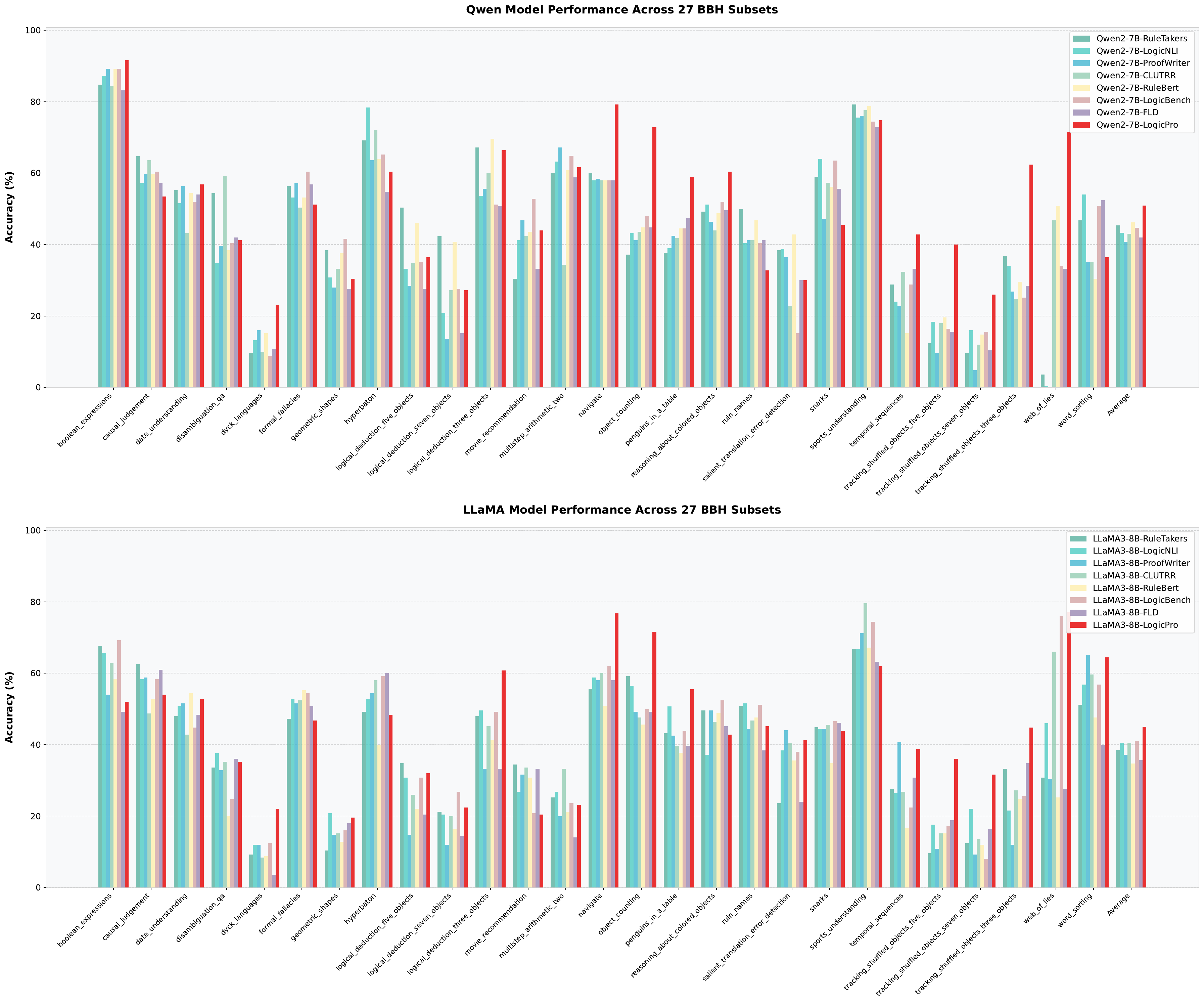}
    \caption{Qwen2-7B and Llama3-8B model performance across 27 BBH subset.}
    \label{fig:BBH_subset}
\end{figure*}

\paragraph{Results for All Subsets of BBH}
In the table \ref{fig:BBH_subset}, we give the histogram results for all baseline data and LogicPro on the 27 subsets of BBH.


\paragraph{Results for Base Data}
In section 3.1, paragraph Experimental design, we mentioned that in order to simulate the actual large model training manufacturers, we choose the generic domain and logical reasoning domain data as the base data. The results of the base data without adding any synthetic data are given in Table \ref{tab:base_data}.
\begin{table*}[htbp]
  \centering
  \resizebox{\linewidth}{!}{%
    \begin{tabular}{lccccccccccc}
    \toprule
    \multicolumn{1}{c}{\textbf{Model }} & \textbf{BBH} & \textbf{LogicBench} & \textbf{DROP} & \textbf{ARLSAT} & \textbf{BoardgameQA} & \textbf{FOLIO} & \textbf{GSM8K} & \textbf{Multi-LogiEval} & \textbf{ProofWriter} & \textbf{MATH} & \textbf{Average} \\
    \midrule
    Qwen2-7B-base & 41.4  & 68.0  & 65.3  & 16.8  & 42.4  & 38.96 & 79.4  & 52.9  & 28.2  & 40.9  & 43.1  \\
    Qwen2-7B-LogicPro & 50.9  & 73.5  & 68.3  & 19.1  & 48.1  & 46.1  & 81.5  & 62.0  & 28.5  & 41.6  & 51.2  \\
    \midrule
    \multicolumn{1}{c}{\textbf{Model }} & \textbf{BBH} & \textbf{LogicBench} & \textbf{DROP} & \textbf{ARLSAT} & \textbf{BoardgameQA} & \textbf{FOLIO} & \textbf{GSM8K} & \textbf{Multi-LogiEval} & \textbf{ProofWriter} & \textbf{MATH} & \textbf{Average} \\
    \midrule
    Llama3-8B-base & 40.5  & 62.3  & 65.3  & 12.4  & 40.0  & 43.73 & 69.7  & 60.1  & 30.8  & 20.0  & 41.6  \\
    Llama3-8B-LogicPro & 45.0  & 67.9  & 68.8  & 15.2  & 44.3  & 48.3  & 74.2  & 68.0  & 37.7  & 23.0  & 46.2  \\
    \bottomrule
    \end{tabular}%
    }
    \caption{Result of not adding any synthetic data (only two types of base data).}
  \label{tab:base_data}%
\end{table*}%

\section{Details of Analysis}

\paragraph{Complete results of the ablation study}
The complete results on all benchmarks of the ablation study are presented in Table \ref{tab:full_ablation}.

\begin{table*}[htbp]
  \centering
  
  \resizebox{\linewidth}{!}{%
    \begin{tabular}{cccccccccccc}
    \toprule
    \textbf{Qwen2-7B} & \textbf{BBH} & \textbf{LogicBench} & \textbf{DROP} & \textbf{AR-LSAT} & \textbf{BoardgameQA} & \textbf{FOLIO} & \textbf{GSM8K} & \textbf{Multi-LogiEval} & \textbf{ProofWriter} & \textbf{MATH} & \textbf{Average} \\
    \midrule
    Source\_LeetCode & 41.0  & 63.4  & 58.4  & 18.3  & 32.8  & 38.2  & 79.7  & 54.8  & 26.2  & 40.6  & 42.2  \\
    LogicPro-w/o.Inter-Var & 48.2  & 69.6  & 62.9  & 16.5  & 47.6  & 50.7  & 80.5  & 62.9  & 30.0  & 40.1  & 49.0  \\
    LogicPro-w.Inter-Var & 50.9  & 73.5  & 68.3  & 19.1  & 48.1  & 46.1  & 81.5  & 62.0  & 28.5  & 41.6  & 51.2  \\
    \midrule
    \textbf{Llama3-8B} & \textbf{BBH} & \textbf{LogicBench} & \textbf{DROP} & \textbf{AR-LSAT} & \textbf{BoardgameQA} & \textbf{FOLIO} & \textbf{GSM8K} & \textbf{Multi-LogiEval} & \textbf{ProofWriter} & \textbf{MATH} & \textbf{Average} \\
    \midrule
    Source\_LeetCode & 36.6  & 51.4  & 62.1  & 12.8  & 38.9  & 48.2  & 69.5  & 62.4  & 31.6  & 20.5  & 38.5  \\
    LogicPro-w/o.Inter-Var & 44.0  & 63.6  & 66.8  & 13.6  & 43.5  & 56.0  & 73.9  & 63.1  & 36.6  & 20.0  & 45.1  \\
    LogicPro-w.Inter-Var & 45.0  & 67.9  & 68.8  & 15.2  & 44.3  & 48.3  & 74.2  & 68.0  & 37.7  & 23.0  & 46.2  \\
    \bottomrule
    \end{tabular}%
    }
    \caption{Complete results of the ablation study.}
  \label{tab:full_ablation}%
\end{table*}%

\paragraph{Complete results of the Performance Gain}

The complete results of Performance Gain on All Baseline Data on all baselines are in Table \ref{tab:full_performance_gain}.
\begin{table*}[htbp]
  \centering
  \resizebox{\linewidth}{!}{%
    \begin{tabular}{cccccccccccc}
    \toprule
    \textbf{Qwen2-7B} & \textbf{BBH} & \textbf{LogicBench} & \textbf{DROP} & \textbf{ARLSAT} & \textbf{BoardgameQA} & \textbf{FOLIO} & \textbf{GSM8K} & \textbf{Multi-LogiEval} & \textbf{ProofWriter} & \textbf{MATH} & \textbf{Average} \\
    \midrule
    All\_Baseline\_Data & 43.1  & 92.8  & 63.6  & 15.2  & 48.4  & 60.5  & 79.2  & 65.7  & 67.9  & 39.9  & 47.1  \\
    All\_Baseline\_Data + LogicPro & 48.8  & 96.3  & 66.8  & 20.4  & 55.4  & 64.2  & 80.6  & 71.1  & 67.9  & 41.0  & 52.3  \\
    \midrule
    \textbf{Llama3-8B} & \textbf{BBH} & \textbf{LogicBench} & \textbf{DROP} & \textbf{AR-LSAT} & \textbf{BoardgameQA} & \textbf{FOLIO} & \textbf{GSM8K} & \textbf{Multi-LogiEval} & \textbf{ProofWriter} & \textbf{MATH} & \textbf{Average} \\
    \midrule
    All\_Baseline\_Data & 42.9  & 94.3  & 66.9  & 17.0  & 48.2  & 64.9  & 78.4  & 74.4  & 69.1  & 39.7  & 47.6  \\
    All\_Baseline\_Data + LogicPro & 47.6  & 95.0  & 68.6  & 16.5  & 51.6  & 61.9  & 79.4  & 76.8  & 66.4  & 40.2  & 51.2  \\
    \bottomrule
    \end{tabular}%
    }
  \caption{Complete results of the Performance Gain.}
    
  \label{tab:full_performance_gain}%
\end{table*}%

\section{Prompts}

\subsection{Complete Prompts for The Three Steps in The Main Methodology}

\begin{figure}[htp]
    \centering
    \includegraphics[width=0.5\textwidth]{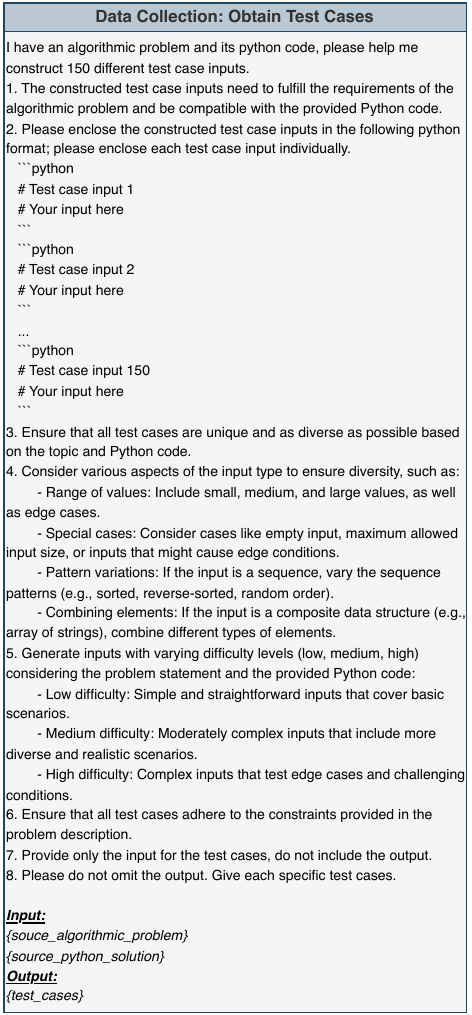}
    \caption{Complete prompts for Data Collection: construct test cases.}
    \label{fig:full_data_collection}
\end{figure}

\begin{figure}[htp]
    \centering
    \includegraphics[width=0.5\textwidth]{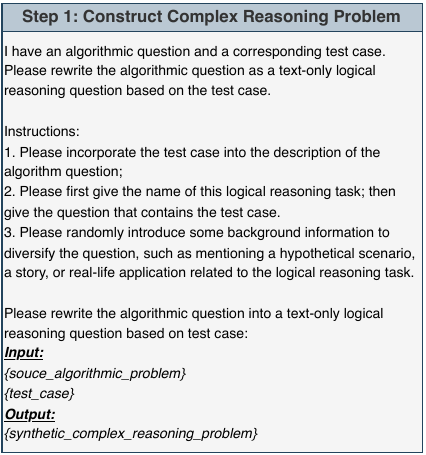}
    \caption{Complete prompts for step 1.}
    \label{fig:full_step_1}
\end{figure}

\begin{figure}[htp]
    \centering
    \includegraphics[width=0.5\textwidth]{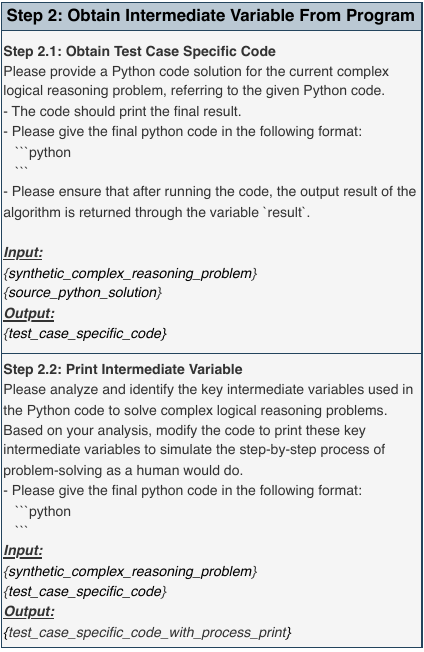}
    \caption{Complete prompts for step 2.}
    \label{fig:full_step_2}
\end{figure}

\begin{figure}[htp]
    \centering
    \includegraphics[width=0.5\textwidth]{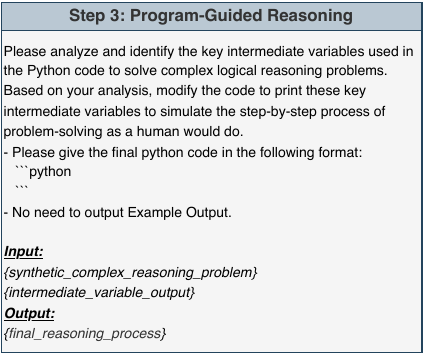}
    \caption{Complete prompts for step 3.}
    \label{fig:full_step_3}
\end{figure}

The complete prompts for data collection and steps 1, 2, and 3 are in Figure \ref{fig:full_data_collection}, \ref{fig:full_step_1}, \ref{fig:full_step_2}, and \ref{fig:full_step_3}.

\subsection{Prompts in the remaining modules}
\label{sec:modules}
\begin{figure}[htp]
    \centering
    \includegraphics[width=0.5\textwidth]{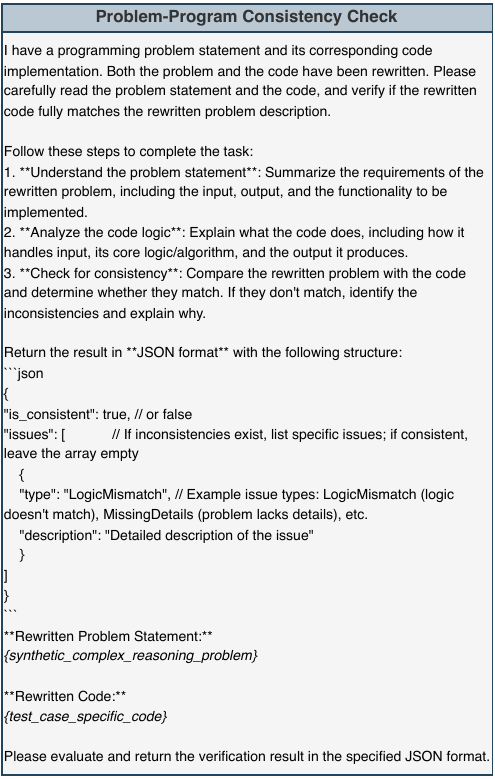}
    \caption{Prompts for Problem-Program Consisitency Check.}
    \label{fig:Consisitency}
\end{figure}

\begin{figure}[htp]
    \centering
    \includegraphics[width=0.5\textwidth]{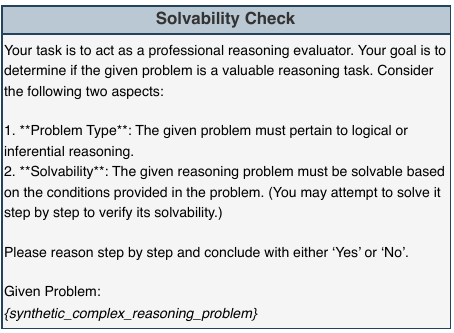}
    \caption{Prompts for Solvability Check.}
    \label{fig:Solvability}
\end{figure}

The complete prompts for Problem-Program Consisitency Check and Solvability Check are in Figure \ref{fig:Consisitency} and \ref{fig:Solvability}.

\begin{figure}[htp]
    \centering
    \includegraphics[width=0.5\textwidth]{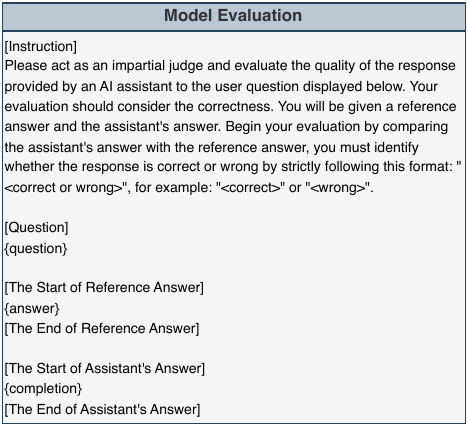}
    \caption{Prompt for Model Evaluation.}
    \label{fig:model_eval}
\end{figure}

In the evaluations, the prompt used for model evaluation is shown in Figure \ref{fig:model_eval}.

\section{LogicPro Examples}
\begin{figure*}[htp]
    \centering
    \includegraphics[width=1.0\textwidth]{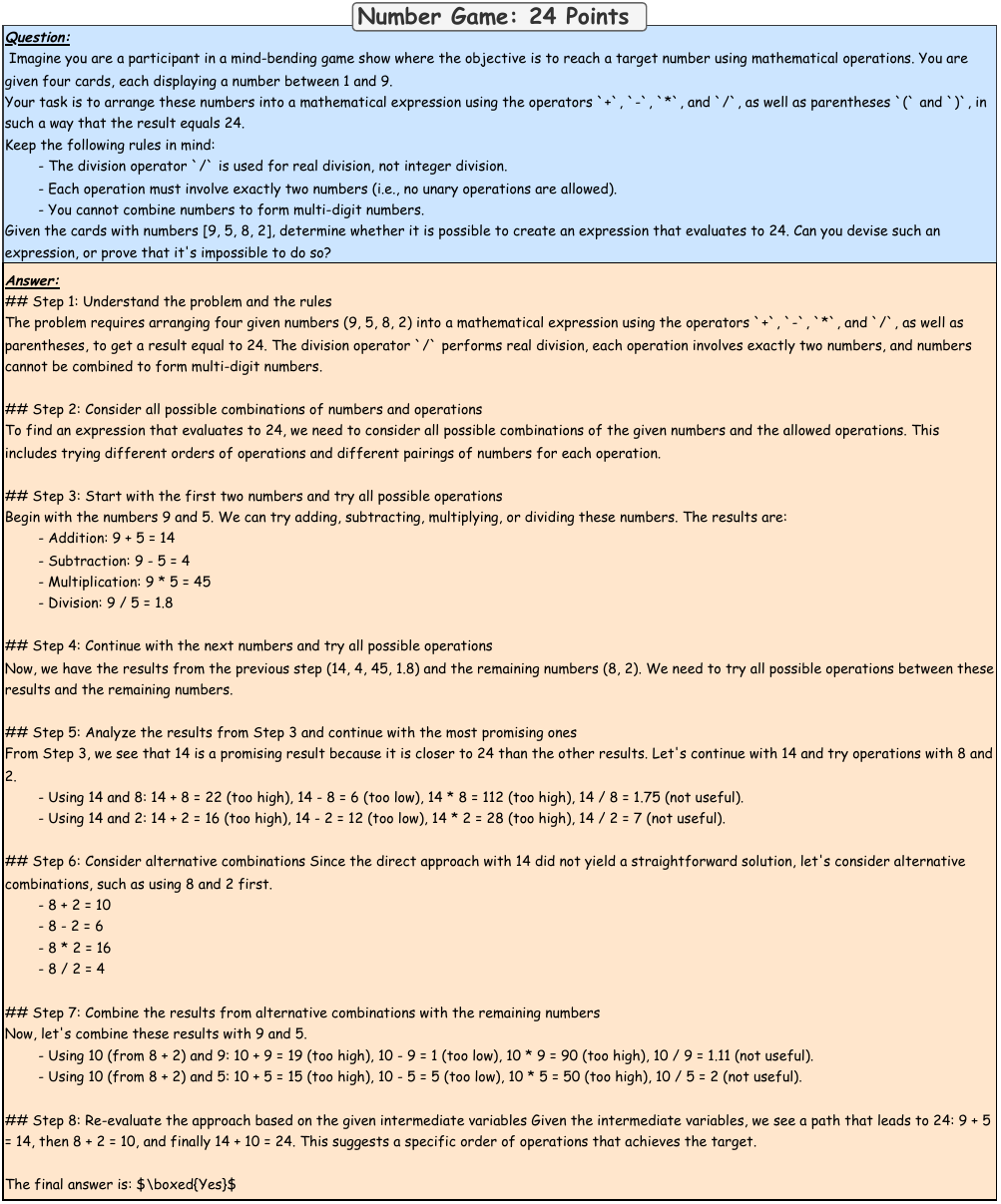}
    \caption{Data Example for 24 Game.}
    \label{fig:24_game}
\end{figure*}

\begin{figure*}[htp]
    \centering
    \includegraphics[width=1.0\textwidth]{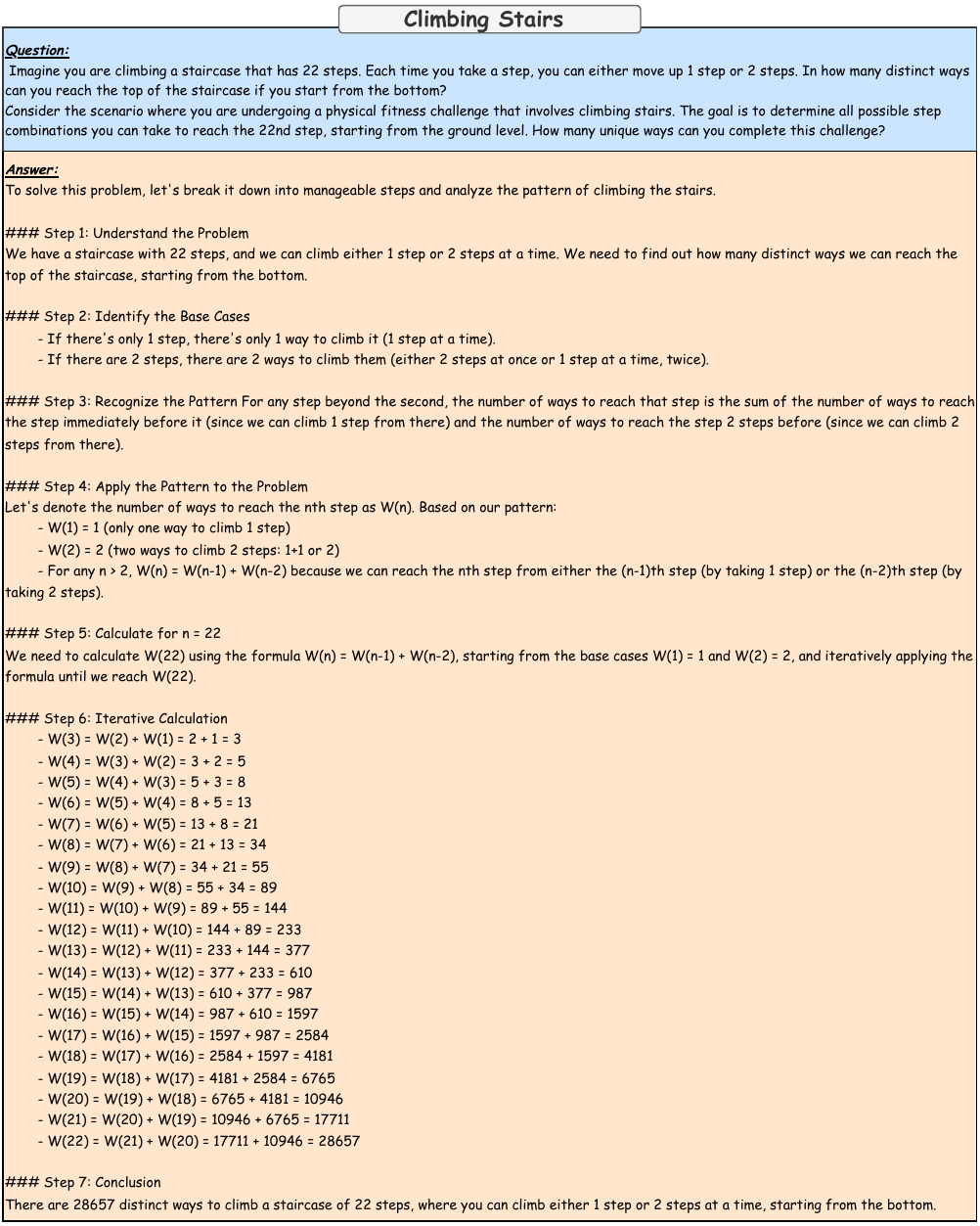}
    \caption{Data Example for Climbing Stairs.}
    \label{fig:climbing_stairs}
\end{figure*}

We give our synthetic data samples in figures \ref{fig:24_game} and \ref{fig:climbing_stairs}, synthesised from two LeetCode algorithm questions, game 24 and stair climbing.


\end{document}